\documentclass[letterpaper, 10 pt, conference]{ieeeconf} 

\IEEEoverridecommandlockouts                              

\usepackage{censor}
\usepackage{graphicx}
\usepackage{booktabs}
\usepackage{amsmath}
\usepackage{tabularx}
\usepackage{array}  
\usepackage{amssymb}
\usepackage{cuted}
\usepackage{listings}
\usepackage{xcolor}
\usepackage{dblfloatfix}
\usepackage{caption}
\usepackage[noadjust]{cite}
\usepackage{caption}
\usepackage{url}
\usepackage{fancyhdr}

\fancypagestyle{arxivnotice}{
    \fancyhf{} 
    \chead{\large\color{gray}This paper has been accepted for publication at the 
    
    IEEE/RSJ International Conference on Intelligent Robots and Systems (IROS), 2026.\copyright\ IEEE}
}

\title{\LARGE 
\vspace{5mm}
\textbf{AeroPlace-Flow: Language-Grounded Object Placement for Aerial Manipulators via Visual Foresight and Object Flow}
}

\author{%
    Sarthak~Mishra$^{*1}$, Rishabh~Dev~Yadav$^{*2}$, Naveen~Nair$^{1}$, Wei~Pan$^{3}$,~and~Spandan~Roy$^{1}$%
    \thanks{This work is supported partly by ``Edge-AI-Di.Vision'' project from Qualcomm Technologies and partly by the `UASAT' project sponsored by MeITY, India. $(^{*})$ denotes equal contribution.}%
    \thanks{$^{1}$Robotics Research Center, IIIT Hyderabad, India.
    Emails: \texttt{sarthak.mishra@research.iiit.ac.in}, \texttt{naveennair2003@gmail.com}, \texttt{spandan.roy@iiit.ac.in}}%
    \thanks{$^{2}$Department of Computer Science, University of Manchester, UK.
    Email: \texttt{rishabh.yadav@postgrad.manchester.ac.uk}}%
    \thanks{$^{3}$Newcastle University, UK
    Email: \texttt{wei.pan2@newcastle.ac.uk}}%
}

\begin{document}

\maketitle
\thispagestyle{arxivnotice}



\begin{strip}
    \centering
    \vspace{-25mm} 
    \includegraphics[width=\textwidth]{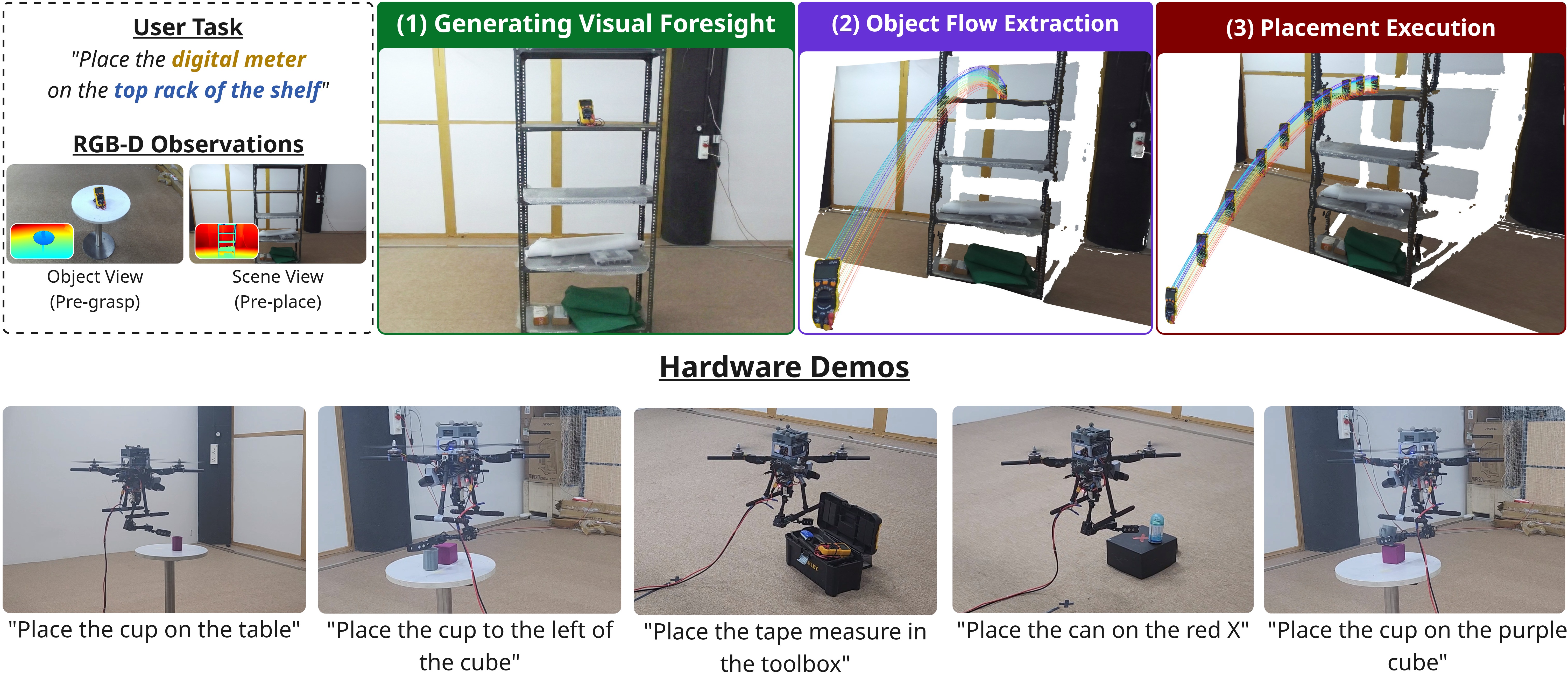}
    \captionof{figure}{\small \textbf{Overview of AeroPlace-Flow.} 
    Given a natural language instruction and RGB-D observations of the object and placement scene, our method infers a collision-free object flow for aerial manipulation in 3 main steps. 
    (1) {Visual Foresight:} A language-conditioned image editing model generates a goal image of the scene with the object placed according to the instruction. 
    (2) {Object Flow Extraction:} The generated image is converted into a metrically consistent 3D scene, contact footprints are estimated, and the original object geometry is used to compute a collision-free object flow trajectory. 
    (3) {Placement Execution:} The aerial manipulator tracks the inferred object flow to execute the placement. 
    Bottom: Hardware demonstrations of language-conditioned aerial placement tasks in diverse scenarios. *Cable connected to drone is only for supplying power.}
    \label{fig:arch}
    \vspace{-4mm} 
\end{strip}

\begin{abstract}
Precise object placement remains underexplored in aerial manipulation, where most systems rely on predefined target coordinates and focus primarily on grasping and control. Specifying exact placement poses, however, is cumbersome in real-world settings, where users naturally communicate goals through language. In this work, we present AeroPlace-Flow, a training-free framework for language-grounded aerial object placement that unifies visual foresight with explicit 3D geometric reasoning and object flow. Given RGB-D observations of the object and the placement scene, along with a natural language instruction, AeroPlace-Flow first synthesizes a task-complete goal image using image editing models. The imagined configuration is then grounded into metric 3D space through depth alignment and object-centric reasoning, enabling the inference of a collision-aware object flow that transports the grasped object to a language and contact-consistent placement configuration. The resulting motion is executed via standard trajectory tracking for an aerial manipulator. AeroPlace-Flow produces executable placement targets without requiring predefined poses or task-specific training. We validate our approach through extensive simulation and real-world experiments, demonstrating reliable language-conditioned placement across diverse aerial scenarios with an average success rate of 75\% on hardware.
\\ \\
{\footnotesize
\noindent Website:  \url{https://aeroplaceflow.github.io/}}
\end{abstract}

\section{Introduction}
Aerial manipulators (AMs) extend the reach of robotics to cluttered, elevated, and hard-to-access environments, enabling physical interaction beyond the workspace limitations of ground robots \cite{ollero2021past}. Prior work such as \cite{Ubellacker24npj-softDrone3, Bauer2022_markerlessGrasp, yadav2025integrated, yadav2024modular} highlight the promise of AMs for inspection, infrastructure maintenance, warehouse logistics, and disaster response. 
While grasping and control have gained more attention, object placement—which ultimately determines task completion—remains comparatively underexplored. Existing aerial manipulation pipelines typically formulate placement as as a two-stage process: manually specifying a 3D pose - \textbf{where}, followed by executing the motion required to realize that pose and coarsely releasing the object - \textbf{how}. Such interfaces require users to provide precise metric coordinates in the reference frame, which is cumbersome and unintuitive.
As aerial systems move toward greater autonomy, the need for more natural interfaces becomes critical. Language offers a scalable and intuitive medium for communicating goals, allowing users to specify tasks such as ``place the object on the rack'' without enumerating exact positions. These observations motivate a focused study of language grounded object placement in aerial settings.

Progress is seen in robotic manipulation to incorporate language via large generative models \cite{saycan,cliport,rt2}. Inspired by human capabilities, prior works \cite{ni2024generate, image_visual_planner, ebert2018visual} demonstrate that generative models, which enjoy capabilities like visual foresight, can guide manipulation through synthesized subgoal images or predictive visual rollouts. In parallel to these image-based approaches, \cite{dharmarajan2025dream2flow, novaflow} leverage generative video models combined with object flow to extract task-relevant motion trajectories and enable continuous manipulation in tasks such as drawer opening or cloth folding. However, transferring these approaches directly to aerial manipulators to solve placement tasks remains challenging and largely unexplored.

In the context of AMs, the primary requirement is to determine \textbf{where} the object should be placed in the scene given a language instruction, after which the system can focus on \textbf{how} to place it. We find that off-the-shelf image editing models offer a practical way to generate visual foresight that is transferrable to AM's without training. Once the desired configuration is inferred, we turn to infer object flow to capture how the object should move through space to reach its target configuration while avoiding collisions. 


Building on these insights, we introduce \textbf{AeroPlace-Flow}, as shown in Fig. \ref{fig:arch},  a framework for language-conditioned object placement with aerial manipulators that combines visual foresight generation with metric object flow inference. Given a language instruction, an observation of the object before grasping, and an observation of the placement scene, AeroPlace-Flow first leverages off-the-shelf image editing models to synthesize a goal image depicting the completed task. This generated image serves as a semantic placement hypothesis that specifies \emph{where} the object should be placed. We recover a metrically consistent 3D scene from the generated image and infer a collision-free object flow that transports the object from its current configuration in the gripper to the desired placement configuration. Finally, the inferred flow is executed by the aerial manipulator using standard trajectory tracking.

We evaluate AeroPlace-Flow on a benchmark of 100 language-conditioned placement tasks spanning tabletop, relative positioning, stacking, and shelf scenarios, and further demonstrate the approach on a real aerial manipulator platform. Experimental results show that language-conditioned visual foresight provides a reliable semantic interface for placement specification, while the proposed object flow inference module recovers accurate metric placement trajectories with a \textbf{success rate of 80\% on the benchmark} and transfer successfully to \textbf{hardware execution with 75\% success rate}.

Our main contributions are summarized as follows:
\begin{itemize}
    \item We propose visual foresight for aerial object placement, enabling off-the-shelf image editing models to synthesize goal scenes from language instructions.
    \item We propose a method to recover collision-free 3D object flow from generated goal images by enforcing metric consistency, estimating contact footprints, and optimizing trajectories in object space.
    \item We construct a benchmark of 100 language-conditioned placement tasks to systematically evaluate visual foresight generation and object flow inference for placement.
    \item We demonstrate real-world aerial manipulation experiments showing that the inferred object flows can be executed on hardware to achieve successful placements.
\end{itemize}

\section{Related Work}

\textbf{Aerial Manipulation and Control:}
Aerial manipulation has advanced in modeling, perception, and control, with key design challenges summarized in \cite{ollero2021past}. Agile grasping with onboard perception has been demonstrated in \cite{Ubellacker24npj-softDrone3,Bauer2022_markerlessGrasp}, following earlier vision-guided aerial interaction systems \cite{Kim2016_visionGuidedAerial}. Foundational quadrotor control methods include \cite{mellinger2011mpc,lee2010geometric}, while learning-augmented MPC and data-driven dynamics compensation improve robustness \cite{torrente2021dataDrivenMPC,kaufmann2020deepDrone}. However, most systems still assume predefined metric placement goals. Residual dynamics learning has reduced dependence on exact analytical models \cite{cao2024computation,das2025dronediffusion,ujjawal2026learn,yadav2026learning,ujjawal2025aermani,yadav2025arcade,yadav2026physics}.

\textbf{Language-Grounded Robotic Manipulation:}
Language provides a natural interface for robotic task specification. SayCan \cite{saycan}, RT-2 \cite{rt2}, and PaLM-E \cite{driess2023palme} combine vision-language models with control, while CLIPort \cite{cliport}, PerAct \cite{peract}, and Diffusion Policy \cite{diffusionpolicy} learn language- and vision-conditioned visuomotor policies. Grasp prediction has also progressed through \cite{mahler2017dexnet,graspnet,contactgraspnet}. These methods primarily target ground manipulators, although recent aerial systems use language and vision for grasping and skill selection \cite{singh2026aerograb,mishra2025aermani, song2025soranav}.

\textbf{Visual Foresight and Generative Subgoal Imagination:}
Visual foresight predicts future observations for control \cite{finn2017deep,ebert2018visual}, while world models learn latent predictive environment representations \cite{ha2018world}. More recent approaches use generative models to synthesize language-conditioned goal images or subgoals \cite{ni2024generate,image_visual_planner}.

\textbf{Object Flow and Learning-Based Placement:}
Dream2Flow \cite{dharmarajan2025dream2flow} and NovaFlow \cite{novaflow} convert generated visual dynamics into object-centric motion. Pick2Place \cite{pick2place} and AnyPlace \cite{anyplace} address task-aware placement and rearrangement for fixed-base manipulators. These methods, however, remain focused on ground manipulation and do not directly address language-conditioned aerial placement.

\section{Methodology}
We first introduce the problem formulation and overview of AeroPlace-Flow in Sec.~III-A. We use off-the-shelf image editing models for generating  visual foresight in Sec.~III-B, then extract metric collision-free object flow in Sec.~III-C, and discuss how to realize this displacement through standard trajectory tracking methods for aerial manipulation in Sec.~III-D.
\subsection{Problem Formulation and Overview}
Given a natural language instruction $L$, RGB-D observations of the object $(I_{obj}, D_{obj})$ and the placement scene $(I_{scene}, D_{scene})$, camera intrinsics $K$, and corresponding camera extrinsics $T_{obj}^{cam}$ and $T_{scene}^{cam}$, the objective is to infer a semantically correct and metrically consistent object flow $\mathbf{P}_{1:T} \in \mathbb{R}^{T \times N \times 3}$, that transports the object from its current gripped configuration to a desired placement configuration, effectively reducing the object placement task to a trajectory tracking problem. 

\textbf{Generating Visual Foresight:} We leverage off-the-shelf language-conditioned image editing models to generate an image of scene with the completed task. The model jointly reasons over the language instruction, object and placement scene observations, producing a goal image where the object is placed according to the instruction.

\textbf{Inferring Object Flow:} We decompose object flow inference into three steps. 
First, we recover a metrically consistent 3D scene for the generated image in a world frame.  
Second, we estimate the contact footprint between the generated object and environment .  
Finally, we swap the generated object's geometry with original object geometry with the help of the inferred contact footprint to remove inconsistencies and use point correspondences to get a linear object path between the object held in the gripper and the placed object. This is optimized with collision and smoothness constraints to obtain a collision-free object flow.  

\textbf{Placement Execution:}  
Given the object flow $\mathbf{P}_{1:T}$, the task is reduced to trajectory tracking problem. The flow specifies the desired time-indexed 3D motion of the object geometry, which is executed by the aerial manipulator while maintaining rigid attachment between the object and gripper. Standard control modules are used to track the requested motion.

\subsection{Generating Visual Foresight}

\begin{figure}[h]
\centering
\includegraphics[width=0.5\textwidth]{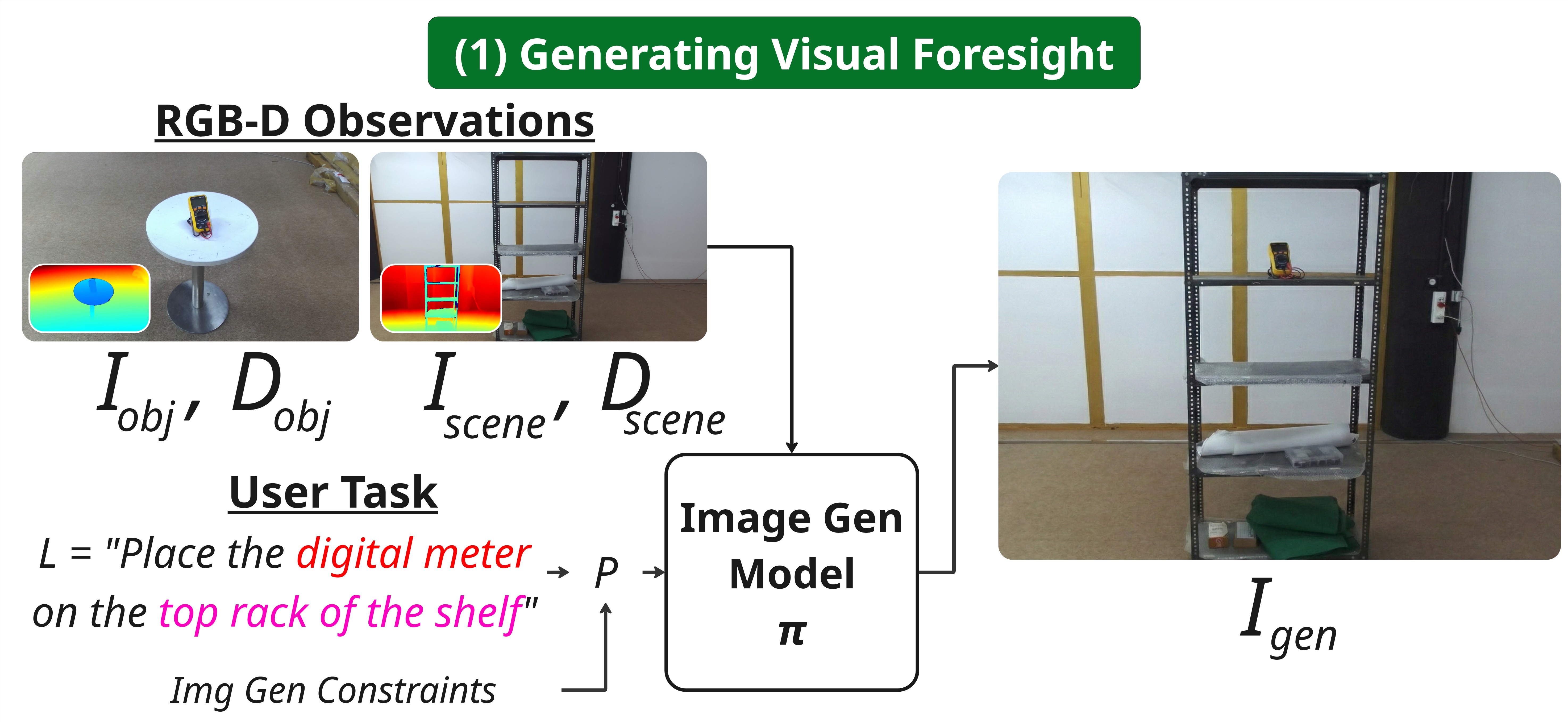}
\caption{\small \textbf{Generating Visual Foresight.} Given RGB-D observations of the object $(I_{obj}, D_{obj})$ and scene $(I_{scene}, D_{scene})$ with a task instruction $L$, an image generation model $\pi$ produces a goal image $I_{\text{gen}}$ depicting the desired final placement.}
\label{fig:imggen}
\end{figure}

Given the language instruction $L$, the object observation $I_{obj}$, and the placement scene observation $I_{scene}$, we use an image editing model $\pi$ with multi-image-context to generate a goal image $I_{gen}$. The generation process as shown in Fig.\ref{fig:imggen}, can be written as:

\begin{equation}
I_{gen} = \pi(I_{obj}, I_{scene}, P),
\end{equation}
where $P$ is a text prompt constructed from the language instruction and task constraints. The prompt $P$ encodes four constraints:  
(1) the object must be placed according to the instruction $L$,  
(2) the image should be generated from the same camera fov as in $I_{scene}$,  
(3) the global scene layout of $I_{scene}$ must remain unchanged except for the added object.  
(4) the object must be placed with an orientation consistent with it appearance in $I_{obj}$. The model jointly reasons over the object appearance from $I_{obj}$ and the spatial layout of $I_{scene}$ to produce a goal image that depicts the completed task. 

As discussed in \cite{pick2place}, the selected grasp strongly influences the set of feasible object placements. In our setting, we do not have explicit feedback about the object’s exact orientation within the gripper at execution time. Therefore, we assume that the object is stably held in an orientation consistent with its observation $I_{obj}$. Constraint (4) operationalizes this assumption by enforcing orientation consistency between the observed object image and the generated goal image.

We observe that image editing models reliably reposition objects when both visual and textual context are sufficiently informative. However, inconsistencies may arise with object scale in $I_{gen}$. We correct this in the next stage, where scale consistency is enforced using metric depth information of the object taken from $D_{obj}$.

\begin{figure*}[h]
\centering
\includegraphics[width=1.0\textwidth]{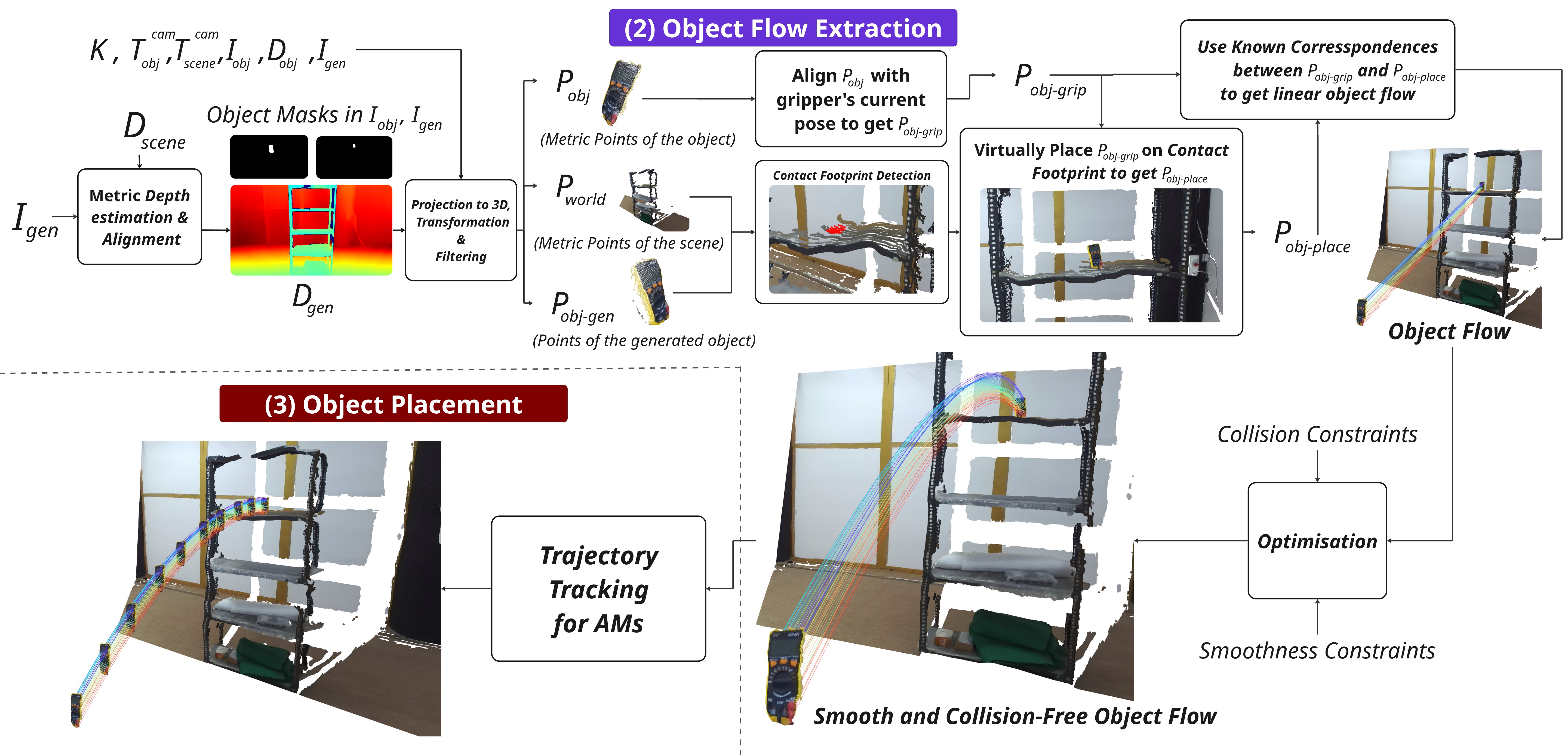}
\caption{\small \textbf{Object flow inference and placement execution.} 
Given the generated goal image $I_{gen}$, we first recover a metrically consistent 3D scene and extract point clouds for the object $\mathcal{P}_{obj}$, generated object $\mathcal{P}_{obj\text{-}gen}$, and world $\mathcal{P}_{world}$. A contact footprint between $\mathcal{P}_{obj\text{-}gen}$ and $\mathcal{P}_{world}$ is estimated to identify the support region. The original object geometry $\mathcal{P}_{obj}$ is then aligned with the current gripper pose and virtually placed on the contact footprint to obtain the desired placement configuration. Known point correspondences between the gripped and placed object are used to generate an initial linear object flow, which is refined through optimization with collision and smoothness constraints to produce a collision-free trajectory. The resulting object flow $\mathbf{P}_{1:T}$ is executed by the aerial manipulator using standard trajectory tracking.}
\label{fig:imgflow}
\end{figure*}

\subsection{Inferring Object Flow}
The complete process is shown in Fig.\ref{fig:imgflow}.\\
\textbf{Metric 3D scene reconstruction:} 
To enable geometric reasoning in $I_{gen}$ with respect to the real world, we recover a metrically consistent depth map for $I_{gen}$ using using monocular depth estimation models as $D_{est}$.
Following \cite{dharmarajan2025dream2flow}, we align $D_{est}$ to the observed scene metric depth $D_{scene}$ using a global scale and shift$(s^*, b^*)$, producing an aligned depth map as: $D_{gen} = (s^* . D_{est} ) + b^*$.

We segment the object in $I_{obj}$ and $I_{gen}$ using image segmentation models to obtain object masks. Using the camera intrinsics $K$ and extrinsics $(T_{obj}^{cam}, T_{scene}^{cam})$, we project the masked depths $D_{obj}$, $D_{gen}$, and $D_{scene}$ into 3D, transform them into the world frame, and apply RANSAC-based filtering to remove projection outliers. This yields a set of three point clouds:

\begin{itemize}
    \item $\mathcal{P}_{obj}$: the metric object geometry reconstructed from $(I_{obj}, D_{obj})$ and its masks.
    
    \item $\mathcal{P}_{obj\text{-}gen}$: the placed object reconstructed from $(I_{gen}, D_{gen})$, providing a semantically correct placement hypothesis but whose geometry may not exactly match the real object.
    
    \item $\mathcal{P}_{world}$: the scene point cloud reconstructed from $(I_{scene}, D_{scene})$, providing the metric geometry of the environment.
\end{itemize}

\textbf{Contact footprint estimation:} 
As mentioned above, the generated object does not preserve the objects geometry and size, therefore directly using $\mathcal{P}_{obj-gen}$ can therefore lead to physically inconsistent placements. However, the generated object is typically positioned at the correct semantic location in the scene. 

We combine this location information with the metric objects geometry in $\mathcal{P}_{obj}$ by first estimating the contact footprint between the world $\mathcal{P}_{world}$ and $\mathcal{P}_{obj-gen}$. The lowest surface points of $\mathcal{P}_{obj-gen}$ are extracted by height-based thresholding and act as contact candidates. For each candidate, distance and tolerance criteria are applied to identify the corresponding support surface in $\mathcal{P}{world}$. The resulting set defines a dense contact footprint in the world frame, representing the support region between the object and world.

\begin{figure*}[t]
\centering
\includegraphics[width=1.0\textwidth]{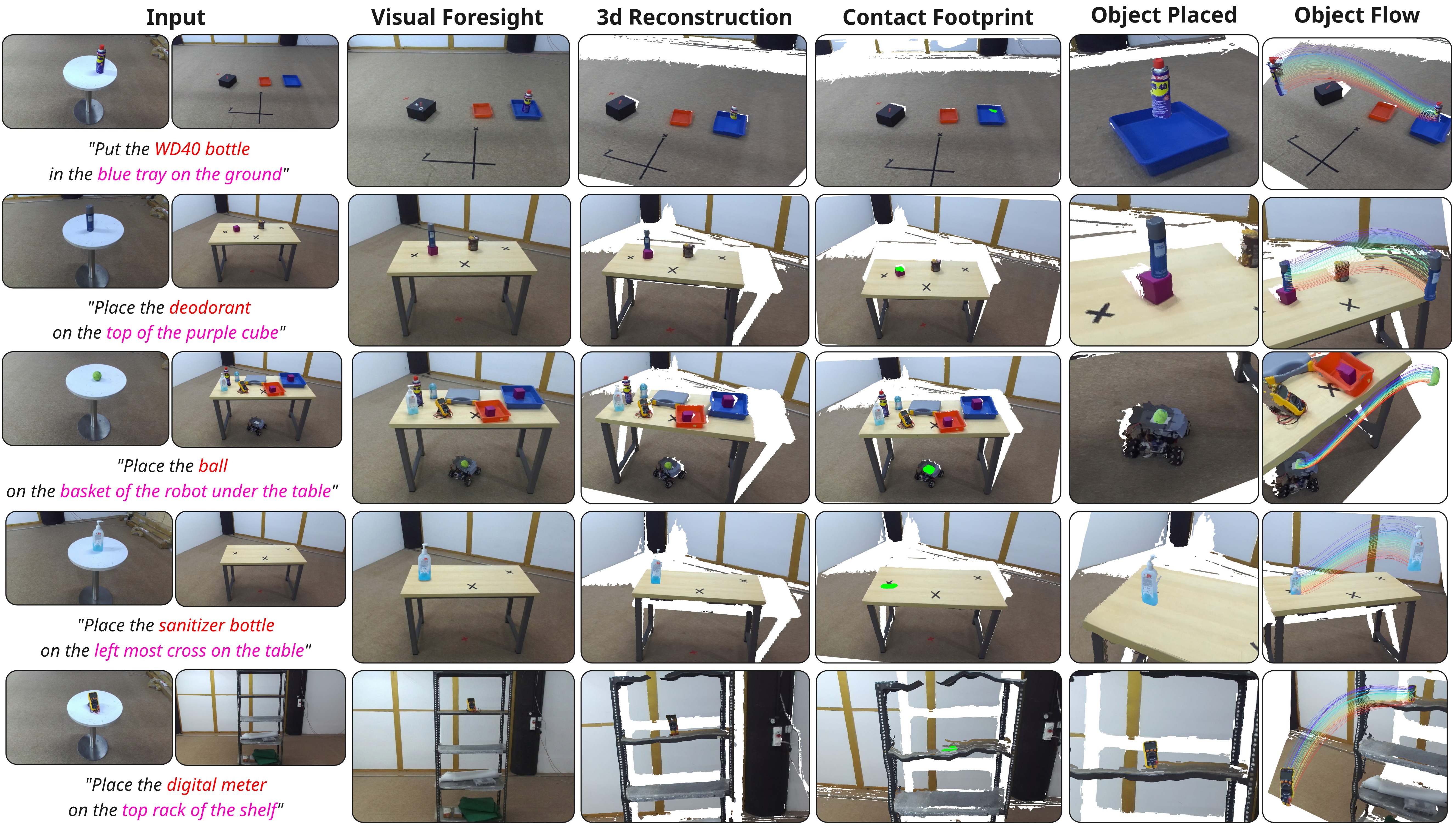}
\caption{\small Representative examples from the 100-task benchmark evaluation. Each row illustrates the full AeroPlace-Flow pipeline for a language-conditioned placement task. From left to right: input object and scene images, generated visual foresight from the image editing model, reconstructed 3D scene, estimated contact footprint(colored in green), resulting placed object pose, and the inferred collision-free object flow trajectory.}
\label{fig:benchtest}
\end{figure*}

\textbf{Object flow computation:} 
If the object consists of $N$ points, the object flow is defined as the time-indexed 3D trajectory of all object points: $\mathbf{P}_{1:T} \in \mathbb{R}^{T \times N \times 3}$,
where each slice $\mathbf{P}_t \in \mathbb{R}^{N \times 3}$ represents the object geometry at time step $t$. We refer to $\mathbf{P}_{1:T}$ as the 3D object flow.

$\mathcal{P}_{obj}$ represents the object kept at a position before grasping, we transform it to the current gripper's position by using the AM's pose and arm joint angles feedback to get $\mathcal{P}_{obj-grip}$. Next, we align and place a copy of $\mathcal{P}_{obj-grip}$ on the contact footprint's center essentially replacing $\mathcal{P}_{obj-gen}$ with $\mathcal{P}_{obj-grip}$.

We now have two rigidly related configurations of identical point sets: one in the gripper and one at the placement location. Since the points are copies, point-wise correspondences are known. We first construct a straight-line interpolation between the two configurations to obtain an initial object flow connecting the start and goal poses.

The linear interpolation provides a coarse displacement but does not account for environmental collisions or motion smoothness. Therefore, we refine this initial path using trajectory optimization with collision avoidance and smoothness constraints. In particular, we employ a sequential convex optimization approach similar to TrajOpt \cite{schulman2014trajopt}, where the object trajectory is iteratively adjusted to maintain clearance from the scene geometry represented by $\mathcal{P}_{world}$ while ensuring smooth motion between consecutive configurations.

The optimized sequence of intermediate configurations transports the object from the current gripper pose to the desired placement configuration while respecting geometric constraints. This sequence forms the final object flow $\mathbf{P}_{1:T}$ used for execution.

\subsection{Placement Execution}
Given the collision-aware object flow $\mathbf{P}_{1:T}$, we extract an aerial manipulation trajectory that realizes the desired transport while respecting platform stability and kinematic feasibility as shown in Fig.\ref{fig:imgflow}. The object flow specifies the desired time-indexed trajectory of all object points in the world frame. Since the object is rigidly attached to the gripper during transport, these point trajectories implicitly determine the motion that the end-effector must follow.

We operate the aerial manipulator in Cartesian end-effector space. For each step in $\mathbf{P}_{1:T}$, a target end-effector configuration is computed and tracked through coordinated motion of the drone base and arm joints. The controller ensures that the commanded end-effector configuration is achieved while maintaining stable flight.

\section{Experimentation}

\begin{table*}[t]
\footnotesize
\renewcommand{\arraystretch}{1.2}
\caption{\small Visual foresight generation performance across 100 language-conditioned placement tasks.}
\centering
\resizebox{\textwidth}{!}{
\begin{tabular}{lccccccccccccccc}
\toprule

& \multicolumn{3}{c}{Nano Banana Pro}
& \multicolumn{3}{c}{Qwen-Image-Edit}
& \multicolumn{3}{c}{FLUX.1 Kontext}
& \multicolumn{3}{c}{GPT-Image}
& \multicolumn{3}{c}{Omni-Gen2} \\

\cmidrule(lr){2-4}
\cmidrule(lr){5-7}
\cmidrule(lr){8-10}
\cmidrule(lr){11-13}
\cmidrule(lr){14-16}

\textbf{Scenario}
& SE$\uparrow$ & IP$\downarrow$ & HA$\downarrow$
& SE$\uparrow$ & IP$\downarrow$ & HA$\downarrow$
& SE$\uparrow$ & IP$\downarrow$ & HA$\downarrow$
& SE$\uparrow$ & IP$\downarrow$ & HA$\downarrow$
& SE$\uparrow$ & IP$\downarrow$ & HA$\downarrow$ \\

\midrule

Tabletop
& 23 & 2 & 0
& 22 & 2 & 1
& 21 & 3 & 1
& 20 & 3 & 2
& 18 & 4 & 3 \\

Shelf
& 22 & 2 & 1
& 21 & 3 & 1
& 20 & 3 & 2
& 18 & 4 & 3
& 16 & 5 & 4 \\

Stacking
& 21 & 3 & 1
& 20 & 3 & 2
& 18 & 4 & 3
& 17 & 4 & 4
& 14 & 6 & 5 \\

Relative
& 22 & 2 & 1
& 21 & 3 & 1
& 19 & 4 & 2
& 18 & 4 & 3
& 15 & 5 & 5 \\

\midrule
\textbf{Overall (100)}
& \textbf{88} & \textbf{9} & \textbf{3}
& \textbf{84} & \textbf{11} & \textbf{5}
& \textbf{78} & \textbf{14} & \textbf{8}
& \textbf{73} & \textbf{15} & \textbf{12}
& \textbf{63} & \textbf{20} & \textbf{17} \\

\midrule
\textbf{Avg Runtime (sec)}
& \multicolumn{3}{c}{12.5}
& \multicolumn{3}{c}{11.5}
& \multicolumn{3}{c}{14.5}
& \multicolumn{3}{c}{11.0}
& \multicolumn{3}{c}{21.0} \\
\bottomrule
\end{tabular}
}
\label{tab:visual_foresight_results}
\end{table*}

We evaluate AeroPlace-Flow to answer the following questions:
1) Can language-conditioned image editing serve as a reliable interface for extracting 3D object flow and object placement?
2) Where do failures occur within the pipeline?
3) How does the choice of generative model affect performance of the pipeline?
4) Does the inferred object flow transfer to successful aerial execution?

In the next sections, we first perform evaluation of the visual foresight generation and object flow inference modules on a custom benchmark, examples of which are shown in Fig.\ref{fig:benchtest}, and then the complete end to end pipeline on a hardware setup and discuss the various results.

\subsection{Implementation Details}
Unless mentioned otherwise, Google Nano Banana Pro model via API is used for generating images, equipped with the following prompt structure: 

\textit{"Place the \_\_\_ from the first image on the \_\_\_ in the second image, generate the image from exactly the same camera angle as in the second image, do not change anything else in the image, place the \_\_\_ with the same orientation as in the 1st image".}

The prompt is edited slightly to add task details such as object and scene name. 
DepthAnythingV3\cite{depthanything3} model is used for monocular depth estimation, SAM3\cite{carion2025sam3segmentconcepts} is used for generating the object relevant masks.

\textbf{Aerial Manipulator Hardware:} A custom made Tarot 650 quadcopter is used for testing, equipped with a CUAV X7+ flight controller running PX4 flight software, ZED depth camera, NVIDIA Jetson nano super as the companion computer. The experiments were performed in a indoor lab with motion capture system. A custom 3DOF 3d-printed manipulator with gripper was attached to the frame of the quadcopter. ROS2 humble with python nodes was used to control the aerial manipulator. A companion laptop is used to prompt the system and run the depth estimation and segmentation models locally.

\subsection{Benchmark and Evaluation}
\begin{figure}[h]
\centering
\includegraphics[width=0.5\textwidth]{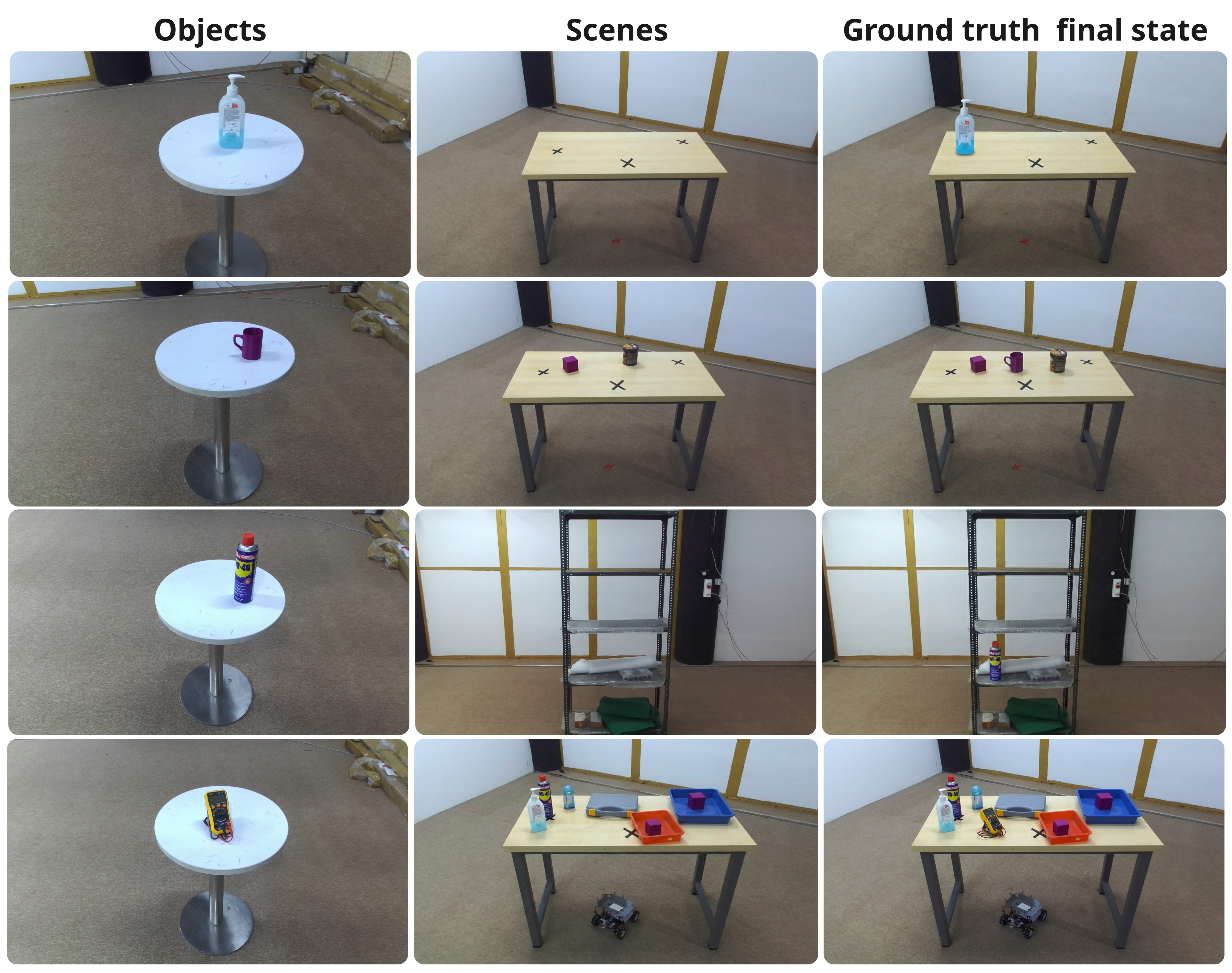}
\caption{\small Overview of the custom benchmark. Each row illustrates a representative task instance with (left) the target object, (middle) the scene configuration, and (right) the corresponding ground-truth final state after placement.}
\label{fig:benchmark}
\end{figure}

We construct a benchmark of 100 language-conditioned object placement tasks to evaluate the visual foresight and object flow inference pipeline. The data is collected, as shown in Fig. \ref{fig:benchmark}, from real-world RGB-D observations in a laboratory environment and covers diverse object categories spanning variations in size, symmetry, and support footprint.
A total of 20 objects and 20 scene images are collected. These are then paired in various combinations to create the 100 tasks.
Tasks are grouped into the following placement classes:

\noindent(i) \emph{Tabletop placement:} placing the object at specified locations on a planar surface (e.g., center, corner, marked region).

\noindent(ii) \emph{Relative positioning:} placing the object next to, in front of, or between reference objects.

\noindent(iii) \emph{Stacking:} placing the object on top of another object.

\noindent(iv) \emph{Shelf placement:} placing the object onto elevated or constrained support regions.

\subsection{Evaluation: Generating Visual Foresight}
We evaluate the Visual Foresight generation module by testing five image editing models on the 100-task benchmark: Google Nano Banana Pro, Qwen-Image-Edit, FLUX.1 Kontext, GPT-Image, Omni-Gen2 via APIs where available and local hosting in others. Each model receives the language instruction $L$, the object observation $I_{obj}$, and the placement scene observation $I_{scene}$. For each scenario and model, we report the following metrics:

\begin{itemize}
     \item \textbf{Image Gen Success (IGS):} Number of Successful Edits where the object from $I_{obj}$ is placed in $I_{scene}$ according to the instruction semantics.
    \item \textbf{Incorrect Placement (IP):} Number of failures due to Incorrect Placement position or orientation.
    \item \textbf{Hallucination (HA):} Number of failures caused by Hallucinations, including introducing new objects, removing existing objects, or altering scene structure.
\end{itemize}

Additionally, we also report the \textbf{Average Runtime} in seconds for the models. Results for this are shown in \ref{tab:visual_foresight_results}.

\subsection{Evaluation: Object Flow Inference}
We evaluate the object flow inference module only for all the successful generations occurred for all the Nano banana model in the previous section which is $88$. Ground truth object poses are available from manual data collection. For each task, we compute the inferred object flow and compare the final placement pose with the ground truth pose. For each scenario, we report the following metrics:

\begin{itemize}
    \item \textbf{Flow Extraction Success (FES):} Percentage of trials where a Valid object Flow is generated and reaches the target support region without collisions.
    \item \textbf{Centroid Pose Error (CPE):} Error between the pose of inferred final object centroid and the ground truth centroid.
\end{itemize}

Results for this are shown in \ref{tab:flow_results}.

\begin{table}[h]
\footnotesize
\renewcommand{\arraystretch}{1.2}
\caption{\small Object flow inference performance evaluated on the $88$ successful visual foresight generations from Nano Banana Pro.}
\centering
\scalebox{1.0}{
\begin{tabular}{lcc}
\toprule
\textbf{Scenario} 
& \textbf{Flow Success $\uparrow$} 
& \textbf{Centroid Pose Error $\downarrow$} \\
\midrule

Tabletop (23) & 22 & 1.8 \\
Shelf (22) & 20 & 2.3 \\
Stacking (21) & 18 & 3.1 \\
Relative (22) & 20 & 2.5 \\

\midrule
\textbf{Overall (88)} & 80 & 2.4 \\

\bottomrule
\end{tabular}
}
\label{tab:flow_results}
\end{table}

\subsection{Evaluation: End-to-End Aerial Execution}
We sample 20 tasks from the benchmark suitable for the lab environment and AM hardware constraints. For the hardware tests, a trial is successful firstly if visual foresight is successful, then the object flow is correctly inferred and finally if the object is placed at the inferred target location within a tolerance of 5cm without collision and remains stably supported after release. The task only proceeds to the next step if the previous one is successful. For each scenario, we report the following metrics:

\begin{itemize}
    \item \textbf{Image Gen Success (IGS):} Number of Successful Edits where the object from $I_{obj}$ is placed in $I_{scene}$ according to the instruction semantics.
    \item \textbf{Flow Extraction Success (FEC):} Number of trials where a valid object flow is generated and reaches the target support region without collisions.
    \item \textbf{Placement Success (PS):} Number of trials where the object is physically placed at the inferred target location within the tolerance without collision and remains stably supported after release.
\end{itemize}

Results for this are shown in \ref{tab:e2e_aerial_results}.
\begin{table}[h]
\footnotesize
\renewcommand{\arraystretch}{1.2}
\caption{\small End-to-end aerial execution results}
\centering
{
\scalebox{1.0}{
\begin{tabular}{lccc}
\toprule
\shortstack{\textbf{Scenario}} & 
\shortstack{\textbf{Image Gen}\\\textbf{Success}} & 
\shortstack{\textbf{Flow Extraction}\\\textbf{Success}} & 
\shortstack{\textbf{Placement}\\\textbf{Success}} \\
\midrule
Tabletop & 5 & 5 & 4 \\
Relative & 5 & 4 & 4 \\
Stacking & 4 & 4 & 4 \\
Shelf & 4 & 4 & 3 \\
\midrule
\textbf{Overall} & \textbf{18/20} & \textbf{17/18} & \textbf{15/20} \\
\bottomrule
\end{tabular}
}}
\label{tab:e2e_aerial_results}
\end{table}

\subsection{Results and Discussion}

\textbf{Can language-conditioned image editing reliably support 3D object flow and placement?}
Table~\ref{tab:visual_foresight_results} shows that modern multi-image editing models generate semantically consistent placement hypotheses across diverse scenarios. Nano Banana Pro performs best, with 88 successful generations out of 100, 9 incorrect placements, and 3 hallucinations. Performance is consistent across tabletop, shelf, stacking, and relative-positioning tasks, although stacking produces slightly more errors because of tighter geometric constraints. These results indicate that language-conditioned visual foresight can preserve object identity and scene structure sufficiently well to provide a reliable semantic prior for geometric reasoning.

\textbf{Where do failures occur within the pipeline?}
Comparing Tables~\ref{tab:visual_foresight_results} and~\ref{tab:flow_results} shows that most failures arise during geometric reconstruction rather than visual foresight. The main source is inaccurate monocular depth under difficult lighting, texture-poor objects, or low object--background contrast, which corrupts the reconstructed 3D geometry. Fewer failures result from incorrect contact-footprint estimation, typically when reconstruction noise causes inaccurate support-region or contact-point identification. Nevertheless, most successful visual hypotheses yield valid object flows, indicating robustness to moderate perception noise.

\textbf{How does the generative model affect performance?}
Table~\ref{tab:visual_foresight_results} shows that successful edits decrease from 88 for Nano Banana Pro to 63 for Omni-Gen2, while incorrect placements and hallucinations increase. Hallucinations are particularly harmful because the geometric pipeline assumes object and scene identity are preserved between the inputs and generated image. Models with stronger multi-image consistency and spatial grounding therefore provide more reliable inputs than models optimized mainly for visual realism.

\textbf{Does inferred object flow transfer to aerial execution?}
Across 20 hardware trials, visual foresight succeeds in 18 cases, object flow extraction in 17 of those cases, and physical placement in 15 cases within a 5\,cm tolerance, yielding a \textbf{75\%} end-to-end success rate. Tabletop and relative-positioning tasks are most reliable, whereas shelf placements are more sensitive to small pose errors because of tighter support regions. Thus, the inferred flows provide sufficiently accurate metric targets for real aerial execution.

\textbf{Summary.}
The results validate language-conditioned visual foresight as an effective intermediate representation for aerial placement. The object-flow module recovers valid placement trajectories in \textbf{80 of 100 benchmark tasks} (80\%), while hardware experiments achieve a \textbf{75\% placement success rate}. Together, visual editing, geometric reconstruction, and trajectory execution enable consistent language-conditioned aerial placement across diverse scenarios.

\section{Conclusion}

We presented \textit{AeroPlace-Flow}, a training-free framework for language-conditioned aerial object placement that combines visual foresight, metric 3D reasoning, and collision-aware object flow inference. Given a language instruction, an image editing model generates a plausible goal scene specifying where the object should be placed, while geometric reconstruction and trajectory optimization determine how to move it safely from the gripper to the target pose.
Evaluation on 100 placement tasks showed that visual foresight provides reliable semantic placement hypotheses and that the proposed grounding module recovers feasible metric object flows in most cases. Hardware experiments further demonstrated effective transfer to aerial execution, achieving a 75\% placement success rate under practical perception and flight constraints.
These results establish visual foresight as a useful intermediate representation for intuitive aerial manipulation without task-specific training or predefined metric goals. Future work will improve robustness in tight-contact settings, incorporate uncertainty-aware perception, and enable closed-loop replanning during flight.

\section*{ACKNOWLEDGMENT}
The authors acknowledge the use of large language models, including Gemini and ChatGPT, solely to improve the manuscript's grammar and linguistic clarity. As image generation is integral to the proposed methodology, selected figures include demonstration images generated using models such as Nano Banana Pro.

\bibliographystyle{IEEEtran}
\bibliography{root}
\end{document}